\documentclass[runningheads]{llncs}

\usepackage[T1]{fontenc}
\usepackage{graphicx}
\usepackage{kotex}
\usepackage{booktabs}
\usepackage{multirow}
\usepackage{bbding}
\usepackage{array}

\newcolumntype{C}[1]{>{\centering\arraybackslash}m{#1}}
\newcolumntype{L}[1]{>{\raggedright\arraybackslash}m{#1}}

\begin{document}

\title{Benchmarking Vendor-Guided Optimizations on Edge NPUs: From NPU Only to End-to-End Performance. Extended technical report}
\titlerunning{Benchmarking Vendor-Guided Optimizations on Edge NPUs}
\author{Youngjin Jo\inst{1,2}\orcidID{0009-0000-8772-5837} \and
Jihwan Kim\inst{1}\orcidID{0009-0000-3446-4805} \and
Minseok Choi\inst{1}\orcidID{0009-0003-6346-4144} \and
Kwanghoon Choi\inst{1}\orcidID{0000-0003-3519-3650}\Envelope \and
Jusung Kang\inst{2}\orcidID{0000-0002-7908-9452} \and
Jaein Kim\inst{2}\orcidID{0009-0006-0739-9052}}
\authorrunning{Y. Jo et al.}
\institute{Chonnam National University, Gwangju, 61186, Korea\\
\email{\{jo.youngjin,213652,213837,kwanghoon.choi\}@jnu.ac.kr}
\and
Electronics and Telecommunications Research Institute, Gwangju, 61012, Korea\\
\email{\{k92492,jaein\}@etri.re.kr}}

\maketitle              
\begin{abstract}
Edge NPU performance depends not only on hardware capabilities, but also on model transformation and the overall deployment pipeline. We propose a benchmarking methodology that separates these effects by comparing three configurations: Baseline, Vendor Guided NPU Only (VG-NPU), and Vendor Guided End-to-End (VG-E2E), where VG-NPU preserves common model I/O and host-side processing while VG-E2E follows the vendor-native deployment pipeline. We apply this methodology to four edge NPUs and three vision workloads. VG-NPU improves both accuracy and Runtime-call latency in 10 of 12 NPU--workload combinations, whereas VG-E2E reduces System-level latency in all combinations but improves accuracy in only four. Moreover, NPU-level performance rankings do not always match End-to-End rankings, and VG-E2E provides an additional 10.4--70.7\% reduction in System-level latency over VG-NPU. These results demonstrate the importance of evaluating vendor-guided optimization separately at the controlled model-transformation and full deployment levels when comparing heterogeneous commercial Edge NPU platforms in practice.

\end{abstract}

%


\section{Introduction}


As AI deployment expands, efficient inference has become increasingly important, particularly in mobile and edge environments with tight power and resource constraints. Neural Processing Units (NPUs) address this demand by accelerating tensor operations commonly used in neural networks, often providing higher efficiency than general-purpose CPUs and GPUs for supported workloads~\cite{Sze2017}. In practice, however, NPU inference performance depends not only on hardware compute capability, but also on the surrounding software stack, including operator support, compiler optimizations, quantization, runtime execution, and input/output processing.


Evaluating NPU-based inference systems requires considering both accuracy and latency. Accuracy reflects how well the predictive quality of the original model is preserved during deployment to an NPU, while latency measures the execution time required to produce the final prediction from an input. In particular, latency can differ substantially depending on whether it measures only NPU inference or end-to-end execution including preprocessing and postprocessing. This study focuses on how accuracy and latency change across the deployment pipelines of different NPUs.


Comparing different NPUs fairly is challenging. NPUs differ not only in hardware architecture and supported operators, but also in their compilers, quantization policies, precision settings, runtime interfaces, and preprocessing/postprocessing pipelines. Consequently, even when starting from the same source model, applying vendor-recommended model transformations and deployment recipes can change not only the final executable model, but also which operations are executed on the NPU and how the overall deployment pipeline is constructed. As a result, comparing only the final end-to-end (E2E) performance reported for each vendor makes it difficult to determine whether an observed performance difference arises from model transformations performed for NPU execution or from the broader deployment pipeline, including preprocessing, postprocessing, and runtime overheads.


Prior work has proposed benchmarking methodologies and evaluation metrics for fair comparison of ML inference systems~\cite{Sze2017,Reddi2020MLPerfInference}, and has analyzed how preprocessing, postprocessing, framework, and runtime overheads beyond model inference affect end-to-end performance~\cite{Millar2025,Buch2021}. Other studies have shown that the execution performance of the same model can vary depending on quantization, compilers, deployment tools, and hardware-specific optimizations~\cite{Hasanpour2025,Ahn2023QuantizationPerformance}. However, these studies have primarily focused on cross-platform comparison, system-level overheads, or the effects of individual optimization techniques. They have not explicitly separated the effect of vendor-guided optimization into two levels: model transformation under fixed model inputs, outputs, and host-side processing, and actual vendor-native end-to-end deployment. By distinguishing these two levels, this study separately examines how model transformation and full deployment affect accuracy and latency.


To investigate this research gap, we define three experimental conditions: Baseline, Vendor Guided NPU Only (VG-NPU), and Vendor Guided End-to-End (VG-E2E). Baseline serves as a reference configuration that permits only the transformations required for NPU execution while limiting optional vendor-specific optimizations. VG-NPU applies the vendor-guided model transformation recipe while preserving the same model inputs and outputs and the same host-side processing as Baseline. VG-E2E applies the full vendor-recommended deployment configuration, including native input/output formats, preprocessing, runtime execution, and postprocessing. With this design, VG-NPU captures the controlled effect of model transformation, whereas VG-E2E measures the performance of the final vendor-guided deployment.

%

Following the proposed evaluation methodology, we analyze three research questions using four NPUs and three vision workloads: ResNet50, YOLOv5m, and YOLOv8s-pose. First, under VG-NPU, both latency and accuracy improve in 10 of the 12 NPU--workload combinations. In contrast, under VG-E2E, System-level latency decreases in all combinations, while accuracy decreases in eight. This result shows that vendor-guided optimization can improve both metrics at the controlled model-transformation level, but that lower latency in the final deployment does not necessarily imply preserved accuracy. 

Second, when comparing three NPUs on the same host, the relative performance ranking among NPUs remains unchanged for YOLOv5m and YOLOv8s-pose, but changes for ResNet50. This indicates that NPU-only performance does not, in general, reliably predict actual end-to-end performance, and that the relationship is workload-dependent. 

Third, VG-E2E achieves lower System-level latency than VG-NPU in all combinations. This suggests that deployment-level processing can provide additional performance gains, while also indicating that such gains should not be attributed to any single optimization component.
Overall, these results show that Edge NPU performance should be evaluated separately at the NPU-level optimization and final deployment levels, rather than being treated as a single performance metric.




The remainder of this paper is organized as follows. Section~\ref{sec:relatedwork} reviews related work. Section~\ref{sec:benchmarkdesign} describes the three experimental configurations, workloads, NPU platforms, and measurement methodology. Section~\ref{sec:evaluationresultsandfindings} presents the main experimental results and discusses the limitations of the study. Finally, Section~\ref{sec:conclusionandfuturework} concludes the paper and outlines directions for future work.

\section{Related Work}
\label{sec:relatedwork}


{\bf Benchmarking methodology and evaluation metrics:}
MLPerf Inference~\cite{Reddi2020MLPerfInference} standardizes workloads, accuracy targets, execution scenarios, and measurement rules to enable fair and reproducible comparisons of ML inference systems across diverse hardware and software platforms. MLPerf Tiny~\cite{Banbury2021MLPerfTiny} extends this benchmarking approach to ultra-low-power TinyML environments by jointly measuring latency, energy, and accuracy. Its modular benchmark design also enables the contributions of different layers of the deployment stack, including models, frameworks, and hardware, to be analyzed separately. These efforts demonstrate that, even across substantially different application domains, fair evaluation of accuracy--latency--energy trade-offs requires considering multiple performance metrics and separating the contributions of individual system layers.

Sze et al.~\cite{Sze2017} similarly emphasize that DNN inference systems should be evaluated using multiple metrics, including accuracy, latency/throughput, energy/power, and hardware cost, and that performance should be measured under representative workloads and runtime conditions. Coleman et al.~\cite{Coleman2017DAWNBench} proposed DAWNBench, highlighting the need for end-to-end system evaluation that captures the combined effects of model architecture, optimization procedures, software frameworks, and hardware.

Following these evaluation principles, this study distinguishes controlled NPU-only execution from native end-to-end deployment to isolate and analyze the effects of vendor-guided optimization.

{\bf NPU/system-level and E2E performance analysis:}
Millar et al.~\cite{Millar2025} compared microcontroller-scale $\mu$NPUs under common workloads and measurement conditions, decomposing execution time into initialization, memory I/O, inference, and post-processing to identify end-to-end performance bottlenecks. 

Buch et al.~\cite{Buch2021} characterized the overheads beyond model inference in mobile AI---including data capture, pre/post-processing, framework, and offloading overheads---as the ``AI tax,'' and analyzed their impact on application-level end-to-end performance. 

Szegedi et al.~\cite{Szegedi2026} separately measured preprocessing, inference, and postprocessing for classification and detection models on edge GPUs, and compared performance across frameworks, implementation languages, and precision settings, demonstrating the need to consider both model inference time and total processing time. 

Kang et al.~\cite{Kang2020} showed that, in accelerator-based visual analytics systems, preprocessing operations such as image decoding and resizing can become throughput bottlenecks that are more significant than model inference itself. 

Ham et al.~\cite{Ham2021}, through their work on NNStreamer, demonstrated that execution configuration can substantially affect the performance of on-device AI pipelines composed of multiple processing stages. By connecting neural network inference and data processing through GStreamer and comparing sequential execution with stream-pipelined execution, they showed that the organization of data flow and parallel execution affects overall latency, throughput, and resource utilization.

While these studies demonstrate the importance of system-level overheads beyond NPU inference, our work distinguishes vendor optimization on commercial Edge NPUs into NPU-only model transformation and full deployment, and compares the performance impact at each level.

{\bf Optimization/toolchain effects:}
EdgeMark~\cite{Hasanpour2025} proposed a benchmarking system that automates model generation, optimization, conversion, and deployment for embedded AI tools, enabling multiple toolchains to be compared under consistent conditions. 

Ahn et al.~\cite{Ahn2023QuantizationPerformance} compared INT8/FP16 and static/dynamic quantization across different frameworks and edge hardware, showing that the effectiveness of the same model optimization can vary depending on software frameworks and hardware support. Baciu et al.~\cite{Baciu2024MLinoBench} similarly compared porting frameworks and quantization configurations on real MCUs, showing that model quantization can reduce model size and classification time at the cost of reduced accuracy.

Ulker et al.~\cite{Ulker2020InferencePerformance} showed that inference performance depends on both the computational structure of a model and framework-specific operator implementations, while MLonMCU~\cite{MLonMCU2024} demonstrated that inference performance and the gains from auto-tuning can vary with TVM schedules and data layouts. 

In contrast to these studies, our work does not focus on comparing individual quantization techniques or deployment tools. Instead, we analyze how complete vendor-recommended optimization recipes affect controlled NPU-only performance and actual end-to-end deployment performance, considering both accuracy and latency.

\section{Benchmark Design}
\label{sec:benchmarkdesign}


To distinguish the effects of vendor-provided NPU optimizations on accuracy and latency between model transformation and final deployment, we define three benchmark configurations: Baseline, Vendor Guided NPU Only (VG-NPU), and Vendor Guided End-to-End (VG-E2E). By comparing Baseline and VG-NPU, we examine the changes introduced by vendor-guided model transformation, while VG-E2E is used to evaluate the final performance of the full vendor-recommended deployment configuration. In Fig.~\ref{fig:benchmark-framework}, we illustrate the three configurations, the execution paths, and the corresponding latency measurement methodology.

\subsection{Benchmark Framework}
\begin{figure*}[h]
    \centering
    \includegraphics[width=\textwidth]{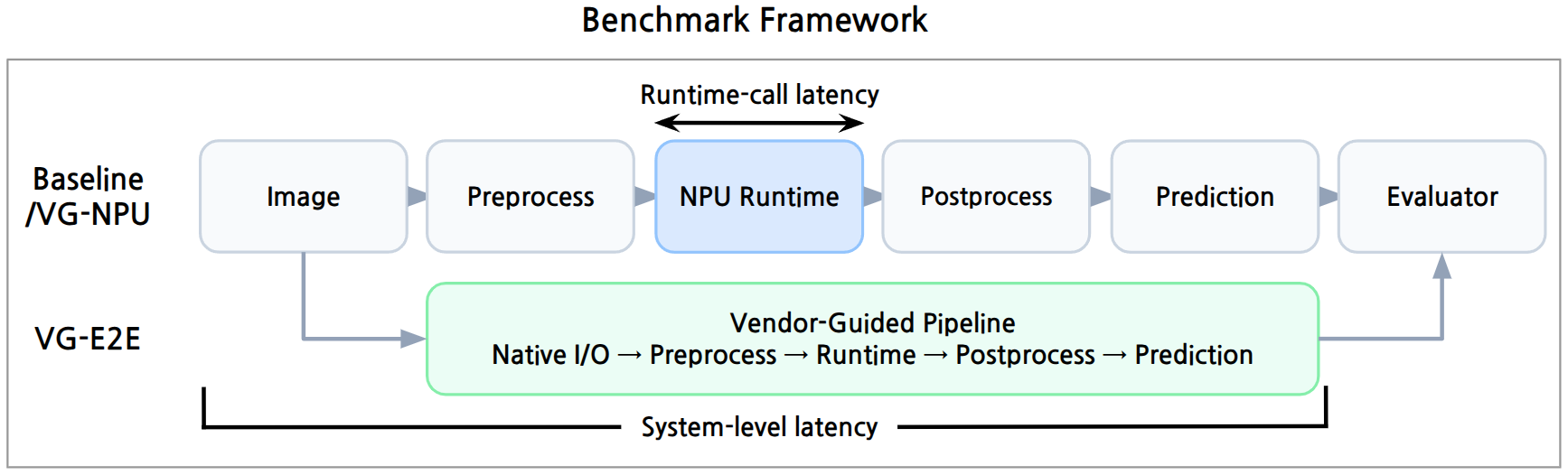}
    \caption{
    Execution paths and latency measurement intervals of the benchmark framework. Baseline and VG-NPU use the same execution path, whereas VG-E2E follows the vendor-guided deployment pipeline. Runtime-call latency measures the elapsed time of the runtime API invocation, which includes NPU execution and any overhead incurred within the runtime call, while System-level latency measures the total execution time from image loading to final prediction.
    }
    \label{fig:benchmark-framework}
\end{figure*}

%

Fig.~\ref{fig:benchmark-framework} illustrates the execution paths of the three benchmark configurations. Baseline and VG-NPU use the same benchmark framework and generate the final prediction through identical preprocessing, NPU runtime invocation, and postprocessing steps. The two configurations are also designed to use the same model inputs and outputs. 
The primary difference between Baseline and VG-NPU is the model transformation configuration used to generate the NPU-executable model.
In contrast, VG-E2E uses the vendor-recommended deployment pipeline, including native input/output formats, preprocessing, runtime, and postprocessing. Accuracy is evaluated using the same evaluator for all configurations.


Runtime-call latency measures the runtime API invocation, while System-level latency measures the time from image loading to final prediction. Since Baseline and VG-NPU use the same runtime interface, their model-transformation effects are compared using Runtime-call latency. Overall deployment performance, including VG-E2E, is compared using System-level latency.


All experiments use synchronous single-stream execution with batch size 1 and the same 5,000-image evaluation set. After 10 warm-up runs, the full set is executed four times. We take the median latency per image across the four runs and report the mean over all 5,000 images.

\subsection{Benchmark Configurations}
%


Table~\ref{tab:benchmark-configurations} summarizes the three benchmark configurations.

\begin{table}
\centering
\caption{Benchmark Configurations}
\label{tab:benchmark-configurations}

\renewcommand{\arraystretch}{1.15}
\begin{tabular}{|C{3.0cm}|C{2.9cm}|C{2.9cm}|C{2.9cm}|}

\hline
\textbf{Category} & \textbf{Baseline} & \textbf{VG-E2E} & \textbf{VG-NPU}  \\
\hline
Source model / weights & Same & Same & Same \\
\hline
Calibration / evaluation & 1,024 / 5,000 images & Same & Same \\
\hline
Batch / execution & 1 / Sync. / Single-stream & Same & Same \\
\hline
Compiler input & Source model subgraph & Source model & Source model subgraph \\
\hline
Quantization & INT 8-bit PTQ & Vendor options & Vendor options \\
\hline
Compiler optimization & Minimal settings & Vendor options & Vendor options \\
\hline
Execution path & Benchmark pipeline & Vendor pipeline & Benchmark pipeline\\
\hline
Latency & Runtime-call / System-level & System-level & Runtime-call / System-level \\
\hline
Evaluation focus & Transformation baseline & Final deployment performance & Model-transformation effect \\
\hline
\end{tabular}
\end{table}



The Baseline configuration serves as the reference point for evaluating the effects of the two vendor-guided optimization configurations. 
\begin{itemize}
\item Model conversion and INT8 post-training quantization are required to make the model executable on each NPU, and these are also applied in the Baseline configuration, since the original FP32 model cannot be executed directly on the NPU without these transformations.

\item  In contrast, vendor-specific optimizations that are not required for model execution are disabled in the Baseline configuration. These include fine-tuning, mixed precision, weight compression, enhanced quantization, graph-level performance optimization, and resource-allocation tuning. These optimizations are enabled only in the vendor-guided configurations, allowing their effects on performance and accuracy to be measured relative to the Baseline.
\end{itemize}



VG-E2E evaluates the deployment performance of the vendor-native execution path, including preprocessing, NPU inference, and postprocessing, using the deployment modules and recommended model-transformation settings provided by each vendor.

\begin{itemize}
\item Model transformation uses the vendor-recommended model graph, quantization, and compiler optimization settings, while execution uses native input/output formats and vendor-specific preprocessing, runtime, and postprocessing.

\item Therefore, changes observed under VG-E2E are interpreted as the effect of the overall vendor-guided deployment rather than that of any single optimization technique.
\end{itemize}


To isolate the effects of NPU inference and model transformation from the overall E2E performance changes, we additionally define the VG-NPU configuration.

\begin{itemize}
\item VG-NPU follows the same execution setup as Baseline, but applies the vendor-recommended model-transformation recipe, including quantization and compiler optimizations.


\item Therefore, the performance difference between Baseline and VG-NPU captures the effect of vendor-recommended model transformation on NPU inference.
\end{itemize}



\subsection{Workloads and Datasets}

Table~\ref{tab:workloads-datasets} summarizes the evaluated workloads and datasets.

\begin{table}
\caption{Evaluation workloads, datasets, and accuracy metrics.}
\label{tab:workloads-datasets}
\centering
\setlength{\tabcolsep}{2.5pt}
\renewcommand{\arraystretch}{1.15}

\begin{tabular}{
|C{0.14\textwidth}
|C{0.12\textwidth}
|C{0.20\textwidth}
|C{0.16\textwidth}
|C{0.19\textwidth}
|C{0.10\textwidth}|}
\hline

\textbf{Model / Task} &
\textbf{Input} &
\textbf{Dataset / Format} &
\textbf{Cal. / Eval.} &
\textbf{Metric} &
\textbf{FP32 Score} \\
\hline

ResNet50 /
Classification &
RGB $224{\times}224$ &
ImageNet-1K /
JPEG + class label &
train 1,024 /
val 5,000 &
Top-1 accuracy &
79.06 \\
\hline

YOLOv5m /
Detection &
RGB $640{\times}640$ &
COCO 2017 /
JPEG + bbox annotation &
train2017 1,024 /
val2017 5,000 &
bbox AP@[.50:.95] &
45.16 \\
\hline

YOLOv8s-pose /
Pose estimation &
RGB $640{\times}640$ &
COCO 2017 /
JPEG + keypoint annotation &
train2017 1,024 /
val2017 5,000 &
keypoint AP@[.50:.95] &
59.39 \\
\hline
\end{tabular}
\vspace{2pt}
\begin{flushleft}
\footnotesize
All accuracy metrics are reported on a 0--100 scale. ResNet50 uses Top-1 accuracy (\%), while YOLOv5m and YOLOv8s-pose use COCO bbox AP and keypoint AP, respectively, multiplied by 100. Bbox AP is averaged over IoU thresholds from 0.50 to 0.95 in increments of 0.05, and keypoint AP is averaged over the corresponding OKS thresholds.
\end{flushleft}
\end{table}



To account for the possibility that NPU optimization effects vary with workload computation and output-processing characteristics, we select ResNet50~\cite{he2016resnet}, YOLOv5m~\cite{jocher2020yolov5}, and YOLOv8s-pose~\cite{jocher2023yolov8} to represent image classification, object detection, and pose estimation, respectively.

Accuracy is evaluated based on the final semantic predictions. For Baseline and VG-NPU, identical model outputs and a common postprocessing pipeline are used so that differences in NPU-specific output tensor layouts do not affect the comparison. For the YOLO workloads, decoding, confidence filtering, and NMS are performed using the same common postprocessor across all NPUs. Detailed output tensor configurations and the threshold settings used for accuracy and latency measurements are provided in Appendix A.

For accuracy evaluation, we use ImageNet-1K~\cite{deng2009imagenet} for ResNet50 and COCO val2017~\cite{lin2014coco} for YOLOv5m and YOLOv8s-pose. Each workload is evaluated on 5,000 images, while calibration uses a fixed set of 1,024 images from the corresponding training dataset. For each workload, the same calibration and evaluation images, in the same order, are used across all NPUs and benchmark configurations to control for dataset-related variation.

\subsection{Vendor-Specific Deployment Configuration}


For each NPU, we apply the optimization and deployment configurations recommended by the corresponding vendor, including quantization, precision, graph transformation, preprocessing/postprocessing, and runtime pipeline.

\begin{itemize}
\item Specifically, Hailo uses optimization/compression-level tuning, mixed precision, and quantization fine-tuning, together with a GStreamer-based execution path and model-integrated normalization or NMS.

\item  DeepX applies compiler optimization levels and enhanced quantization, and uses a DXSTREAM-based pipeline.

\item Mobilint applies compiler optimization, percentile-based quantization, and Hadamard transformation, using a Model Zoo- and \texttt{qbruntime}-based execution setup. 
\end{itemize}

 In VG-NPU, we primarily apply quantization and compiler-level model transformations from these vendor settings, whereas VG-E2E additionally includes vendor-native preprocessing, model-integrated processing, postprocessing, and runtime pipelines. Detailed model-specific options and deployment configurations are summarized in Appendices B and C.

\subsection{NPU Hardware and Software Environments}

We selected four commercially available Edge NPUs from multiple vendors that provide dedicated compiler, runtime, and deployment stacks required to evaluate both the VG-NPU and VG-E2E configurations. Table~\ref{tab:npu-environments} summarizes the hardware and software environments used for each platform.

\begin{table}
\caption{Evaluation NPU Platforms and Software Stacks}
\label{tab:npu-environments}
\centering

\setlength{\tabcolsep}{2pt}
\renewcommand{\arraystretch}{1.2}

\begin{tabular}{
|C{0.12\textwidth}
|C{0.09\textwidth}
|C{0.23\textwidth}
|C{0.10\textwidth}
|C{0.15\textwidth}
|C{0.16\textwidth}|}
\hline

\textbf{Target} &
\textbf{INT8 TOPS}&
\textbf{Host / OS} &
\textbf{CPU Freq.} &
\textbf{Compiler} &
\textbf{Runtime} \\
\hline

Hailo-8 &
26&
Jetson Orin Nano /
Ubuntu 22.04, JetPack 6 &
1.728 GHz &
DFC 3.34 &
HailoRT 4.24 \\
\hline

Hailo-10H &
20&
Jetson Orin Nano /
Ubuntu 22.04, JetPack 6 &
1.728 GHz &
DFC 5.3 &
HailoRT 5.3 \\
\hline

DX-M1 &
25&
Jetson Orin Nano /
Ubuntu 22.04, JetPack 6 &
1.728 GHz &
DX-COM 2.3-rc.5&
DX-RT 3.4 \\
\hline

Regulus &
10&
Regulus SOM /
Vendor Yocto Linux &
1.2 GHz &
QBCompiler 1.1.2 &
qbruntime 1.2 \\
\hline

\end{tabular}
\end{table}


All NPUs are evaluated using the hardware and software stacks provided by their respective vendors. We evaluate Hailo-8 and Hailo-10H from Hailo, DX-M1 from DeepX, and Regulus from Mobilint. Hailo-8, Hailo-10H, and DX-M1 are measured on the same Jetson Orin Nano host, while Regulus is evaluated on the vendor-provided Regulus SOM. To reduce run-to-run variation caused by host-side frequency scaling, the CPU frequency is fixed at 1.728 GHz on the Jetson Orin Nano and 1.2 GHz on the Regulus System on Module (SOM), respectively.







\section{Evaluation Results and Findings}
\label{sec:evaluationresultsandfindings}


In this evaluation, we examine how vendor-guided optimization affects latency and accuracy, and how NPU-only performance translates into end-to-end performance. Table~\ref{tab:accuracy-all-conditions} summarizes the accuracy measured under Baseline, VG-NPU, and VG-E2E, while Table~\ref{tab:e2e-all-conditions} reports the System-level latency for the three configurations. Table~\ref{tab:runtime-call-common-vgn} compares the Runtime-call latency of Baseline and VG-NPU.

\begin{table}[h]
\centering
\caption{Accuracy Results on a 0--100 Scale}
\label{tab:accuracy-all-conditions}

\setlength{\tabcolsep}{5pt}
\renewcommand{\arraystretch}{1.1}

\begin{tabular}{|c|l|r|r|r|}
\hline

\multicolumn{1}{|c|}{\textbf{NPU}} &
\multicolumn{1}{c|}{\textbf{Model}} &
\multicolumn{1}{c|}{\textbf{Baseline}} &
\multicolumn{1}{c|}{\textbf{VG-NPU}} &
\multicolumn{1}{c|}{\textbf{VG-E2E}} \\
\hline

\multirow{3}{*}{Hailo-8}
& ResNet50      & 78.38   & 78.88   & 78.08 \\
& YOLOv5m       & 41.68& 43.33& 39.21\\
& YOLOv8s-pose  & 49.67& 51.68& 48.29\\
\hline

\multirow{3}{*}{Hailo-10H}
& ResNet50      & 78.32   & 78.92   & 77.82 \\
& YOLOv5m       & 42.39& 43.46& 42.80\\
& YOLOv8s-pose  & 50.58& 52.49& 47.84\\
\hline

\multirow{3}{*}{DX-M1}
& ResNet50      & 78.86   & 78.66   & 78.94 \\
& YOLOv5m       & 44.75& 44.82& 43.69\\
& YOLOv8s-pose  & 58.47& 58.90& 58.54\\
\hline

\multirow{3}{*}{Regulus}
& ResNet50      & 78.74   & 78.96   & 77.42 \\
& YOLOv5m       & 43.90& 42.89& 43.38\\
& YOLOv8s-pose  & 56.25& 56.57& 56.66\\
\hline
\end{tabular}%
\end{table}

\begin{table}[h]
\centering
\caption{System-level Latency (ms).}
\label{tab:e2e-all-conditions}

\setlength{\tabcolsep}{5pt}
\renewcommand{\arraystretch}{1.1}

\begin{tabular}{|c|l|r|r|r|}
\hline

\multicolumn{1}{|c|}{\textbf{NPU}} &
\multicolumn{1}{c|}{\textbf{Model}} &
\multicolumn{1}{c|}{\textbf{Baseline}} &
\multicolumn{1}{c|}{\textbf{VG-NPU}} &
\multicolumn{1}{c|}{\textbf{VG-E2E}} \\
\hline

\multirow{3}{*}{Hailo-8}
& ResNet50
& 11.100
& 9.786
& 8.766 \\
& YOLOv5m
& 59.308
& 43.106
& 28.101 \\
& YOLOv8s-pose
& 43.938
& 38.228
& 22.167 \\
\hline

\multirow{3}{*}{Hailo-10H}
& ResNet50
& 16.145
& 15.082
& 9.712 \\
& YOLOv5m
& 124.871
& 114.839
& 33.697 \\
& YOLOv8s-pose
& 92.837
& 89.904
& 32.683 \\
\hline

\multirow{3}{*}{DX-M1}
& ResNet50
& 17.739
& 17.237
& 7.243 \\
& YOLOv5m
& 122.177
& 121.915
& 53.768 \\
& YOLOv8s-pose
& 106.322
& 105.563
& 55.149 \\
\hline

\multirow{3}{*}{Regulus}
& ResNet50
& 34.217
& 34.233
& 28.187 \\
& YOLOv5m
& 122.216
& 122.332
& 96.966 \\
& YOLOv8s-pose
& 193.581
& 192.910
& 86.611 \\
\hline

\end{tabular}%
\end{table}

\begin{table}[h]
\centering
\caption{Runtime-call Latency under the Baseline and VG-NPU (ms).}
\label{tab:runtime-call-common-vgn}

\setlength{\tabcolsep}{5pt}
\renewcommand{\arraystretch}{1.1}

\begin{tabular}{|c|l|r|r|}
\hline

\multicolumn{1}{|c|}{\textbf{NPU}} &
\multicolumn{1}{c|}{\textbf{Model}} &
\multicolumn{1}{c|}{\textbf{Baseline}} &
\multicolumn{1}{c|}{\textbf{VG-NPU}} \\
\hline

\multirow{3}{*}{Hailo-8}
& ResNet50
& 5.506
& 4.187 \\
& YOLOv5m
& 39.640
& 23.271 \\
& YOLOv8s-pose
& 18.656
& 12.891 \\
\hline

\multirow{3}{*}{Hailo-10H}
& ResNet50
& 10.512
& 9.462 \\
& YOLOv5m
& 105.090
& 95.024 \\
& YOLOv8s-pose
& 67.445
& 65.157 \\
\hline

\multirow{3}{*}{DX-M1}
& ResNet50
& 11.862
& 11.449 \\
& YOLOv5m
& 97.082
& 96.851 \\
& YOLOv8s-pose
& 83.571
& 82.892 \\
\hline

\multirow{3}{*}{Regulus}
& ResNet50
& 5.270
& 5.269 \\
& YOLOv5m
& 28.558
& 28.568 \\
& YOLOv8s-pose
& 19.409
& 18.822 \\
\hline

\end{tabular}
\end{table}

\subsection{RQ1: Do vendor-guided optimizations improve both latency and accuracy?}



Comparing Baseline and VG-NPU, accuracy improves in 10 of the 12 NPU--workload combinations, while Runtime-call latency decreases in 11. Both metrics improve simultaneously in 10 combinations, indicating that vendor-guided model transformation can generally reduce Runtime-call latency while preserving or slightly improving accuracy under fixed model inputs, outputs, and host-side processing.
However, accuracy decreases for YOLOv5m on Regulus, and several NPU--workload combinations show only marginal changes in Runtime-call latency, indicating that the benefits are not uniform across all cases.


In contrast, under VG-E2E, System-level latency decreases relative to Baseline in all 12 NPU--workload combinations, whereas accuracy improves in only four and decreases in eight. Thus, vendor-guided deployment consistently improves latency, but does not necessarily improve accuracy. In other words, full-deployment optimization can achieve lower latency at the cost of accuracy degradation for some workloads.

\subsection{RQ2: Does NPU-only performance predict end-to-end performance?}



To examine whether NPU-only performance can predict actual end-to-end performance, we compare the relative performance rankings among NPUs using VG-NPU Runtime-call latency and VG-E2E System-level latency. To minimize the effect of host differences, we focus on Hailo-8, Hailo-10H, and DX-M1, all measured on the same Jetson Orin Nano host. Regulus is excluded from this comparison because it was measured on a separate vendor-provided Regulus SOM.


For ResNet50, the performance ranking differs between NPU-only and end-to-end measurements. Under VG-NPU, Runtime-call latency follows Hailo-8 $<$ Hailo-10H $<$ DX-M1, whereas under VG-E2E, System-level latency follows DX-M1 $<$ Hailo-8 $<$ Hailo-10H. Notably, DX-M1, which is the slowest in the NPU-only measurement, achieves the lowest System-level latency under VG-E2E. This shows that NPU inference performance alone does not always predict overall deployment performance.


In contrast, for YOLOv5m and YOLOv8s-pose, the relative ranking among NPUs remains unchanged across the two measurements. For both workloads, the ranking is Hailo-8 $<$ Hailo-10H $<$ DX-M1 for both VG-NPU Runtime-call latency and VG-E2E System-level latency. These results show that the relationship between NPU-only and end-to-end performance is workload-dependent, and that overall system performance can be influenced not only by NPU inference but also by preprocessing, postprocessing, and vendor-specific execution pipelines.

\subsection{RQ3: How much additional benefit comes from vendor-guided deployment?}



To examine whether vendor-guided deployment provides additional performance gains beyond NPU-level optimization, we compare the System-level latency of VG-NPU and VG-E2E. VG-E2E achieves lower latency than VG-NPU in all 12 NPU--workload combinations, with reductions ranging from 10.4\% to 70.7\%. Particularly large additional improvements are observed for YOLOv5m and YOLOv8s-pose on Hailo-10H, all three workloads on DX-M1, and YOLOv8s-pose on Regulus.


%

VG-NPU and VG-E2E both apply vendor-guided model transformations, but VG-E2E uses a broader vendor-native deployment configuration, including native input/output formats, preprocessing, postprocessing, and runtime pipelines. Since VG-E2E achieves lower System-level latency than VG-NPU in all combinations, the additional latency gains reflect the combined effects of these broader deployment-level changes beyond the controlled VG-NPU configuration.
This suggests that deployment-level optimization can play an important role in further reducing overall latency. However, because multiple components are changed simultaneously, the observed gains cannot be isolated as the effect of any single model transformation, preprocessing, postprocessing, or runtime optimization.

\subsection{Limitations}




This study is limited to the four Edge NPUs, three vision workloads, specific software stacks, and synchronous batch-1 execution, and does not explore all compiler and quantization configurations. In Baseline and VG-NPU, some full-graph optimizations are restricted to preserve identical model inputs, outputs, and host-side processing. In VG-E2E, multiple deployment components change simultaneously, preventing us from isolating the effect of individual optimizations. Because some platforms use different hosts, System-level latency cannot be attributed solely to NPU performance. Power and energy-efficiency evaluation was excluded because the available power telemetry interfaces and measurement coverage differed across platforms, making it difficult to establish a consistent power measurement scope. Consequently, we cannot determine whether the observed latency improvements translate into better power efficiency or lower energy per inference. Thus, our results should be interpreted as the observed effects in vendor-guided optimization within the scope of our experiments, rather than as maximum achievable performance or universal superiority.

\section{Conclusion and Future Work}
\label{sec:conclusionandfuturework}

Using four commercial Edge NPUs and three widely used vision workloads spanning image classification, object detection, and pose estimation, our findings show that Edge NPU performance should be interpreted by distinguishing NPU-level model transformation from final deployment, rather than treating inference performance as a single metric. First, vendor-guided model transformation generally reduces Runtime-call latency while preserving accuracy under fixed model inputs, outputs, and host-side processing. Second, such NPU-level improvements do not always translate directly into final End-to-End performance; for some workloads, the relative performance ranking among NPUs changes after full deployment. Third, vendor-native deployment provides additional System-level latency reductions across all evaluated combinations, but these performance gains do not always preserve accuracy. These findings suggest that comparing Edge NPUs requires considering the latency--accuracy trade-off jointly shaped by model transformation, preprocessing, postprocessing, and runtime pipelines, rather than NPU execution alone. 

Future work will extend the evaluation to a broader range of NPUs and workloads, isolate the effects of individual optimizations, and incorporate workload-specific accuracy constraints to better understand the sources of these trade-offs.



%

\appendix 

\section{Detailed Model I/O and Evaluation Settings}
\label{app:model-io-eval}

Table~\ref{tab:model-io-eval-settings} summarizes the common model output and postprocessing settings used in the Baseline and VG-NPU configurations, together with the threshold settings used for accuracy and latency evaluation. VG-E2E-specific input/output and postprocessing changes are described separately in the vendor-specific deployment configurations.

\begin{table}[h]
\centering
\caption{Model output and evaluation settings.}
\label{tab:model-io-eval-settings}

\setlength{\tabcolsep}{3pt}
\renewcommand{\arraystretch}{1.15}
\footnotesize

\begin{tabular}{
|L{1.75cm}
|L{2.45cm}
|L{2.35cm}
|L{2.10cm}
|L{2.10cm}|}
\hline

\multicolumn{1}{|c|}{\textbf{Model}} &
\multicolumn{1}{c|}{\textbf{Output}} &
\multicolumn{1}{c|}{\textbf{Postprocessing}} &
\multicolumn{1}{c|}{\textbf{Accuracy}} &
\multicolumn{1}{c|}{\textbf{Latency}} \\
\hline

ResNet50
& 1,000 class scores
& Highest-score class
& Top-1 accuracy
& Same prediction procedure \\
\hline

YOLOv5m
& 3 detection heads
& BBox decoding, confidence filtering, and NMS
& conf.\ 0.001; IoU 0.60; max det.\ 300
& conf.\ 0.25; IoU 0.45; max det.\ 1000 \\
\hline

YOLOv8s-pose
& 9 tensors for bbox, class, and keypoints
& BBox/keypoint decoding, confidence filtering, and NMS
& conf.\ 0.001; IoU 0.70; max det.\ 300
& conf.\ 0.25; IoU 0.70; max det.\ 300 \\
\hline

\end{tabular}
\end{table}

For ResNet50, the model produces scores for the 1,000 ImageNet classes,
and the class with the highest score is used as the final prediction.

For YOLOv5m, the Baseline and VG-NPU configurations use the outputs of the 3 detection heads as the common model output. Bounding boxes are decoded into the original image coordinate system, followed by confidence filtering and non-maximum suppression (NMS). The accuracy evaluation follows the validation settings of the official YOLOv5 v7.0 implementation, whereas latency evaluation follows its default prediction settings.

For YOLOv8s-pose, the common model output consists of 9 tensors corresponding to bounding-box, class-probability, and keypoint outputs over 3 scales. Bounding boxes and keypoints are decoded into the original image coordinate system, followed by confidence filtering and
NMS. The accuracy and latency settings follow the validation and prediction defaults, respectively, of Ultralytics v8.4.24.

The lower confidence thresholds used for accuracy evaluation retain low-confidence predictions required for COCO AP computation, whereas latency evaluation uses the default prediction settings of the corresponding reference implementation. The same common postprocessing implementation is used across NPUs for Baseline and VG-NPU.

\section{Vendor-Specific Optimization Options}
\label{app:optimization-options}

Table~\ref{tab:vendor-optimization-options} summarizes the main vendor-specific optimization options applied under the three benchmark configurations. The option names are preserved as exposed by the corresponding vendor toolchains. Model-specific precision and quantization settings are listed separately where the applied configuration differs across workloads.

\begin{table}[h]
\centering
\caption{Vendor-specific optimization changes across configurations.}
\label{tab:vendor-optimization-options}

\setlength{\tabcolsep}{2pt}
\renewcommand{\arraystretch}{1.15}
\footnotesize

\begin{tabular}{
|C{1.45cm}
|L{2.95cm}
|L{3.40cm}
|L{3.40cm}|}
\hline

\multicolumn{1}{|c|}{\textbf{NPU}} &
\multicolumn{1}{c|}{\textbf{Option / Setting}} &
\multicolumn{1}{c|}{\textbf{Baseline vs. VG-NPU}} &
\multicolumn{1}{c|}{\textbf{VG-NPU vs. VG-E2E}} \\
\hline

\multirow{5}{*}[-5ex]{Hailo-8}
& optimization\_level
& $-100 \rightarrow 2$
& Same as VG-NPU \\
\cline{2-4}

& compression\_level
& $0 \rightarrow 1$
& Same as VG-NPU \\
\cline{2-4}

& ResNet50
& W8A8 $\rightarrow$ W4 $\times$ 3 + bias correction
& Same as VG-NPU \\
\cline{2-4}

& YOLOv5m
& W8A8 $\rightarrow$ W4 $\times$ 3 + finetuning
& Same as VG-NPU \\
\cline{2-4}

& YOLOv8s-pose
& W8A8 $\rightarrow$ keypoint A16/W16; 

finetuning off
& finetuning on \\
\hline

\multirow{5}{*}[-5ex]{Hailo-10H}
& optimization\_level
& $-100 \rightarrow 2$
& Same as VG-NPU \\
\cline{2-4}

& compression\_level
& $0 \rightarrow 1$
& Same as VG-NPU \\
\cline{2-4}

& ResNet50
& W8A8 $\rightarrow$ W4 $\times$ 3 + bias correction
& Same as VG-NPU \\
\cline{2-4}

& YOLOv5m setting
& W8A8 $\rightarrow$ W4 $\times$ 4 + finetuning;
  \texttt{conv38} forced to W8
& W4 $\times$ 3 + finetuning;
  Model Zoo allocator recipe \\
\cline{2-4}

& YOLOv8s-pose
& W8A8 $\rightarrow$ keypoint A16/W16 + W4 $\times$ 3
& W4 off; finetuning on \\
\hline

\multirow{2}{*}[-2ex]{DX-M1}
& opt\_level
& $0 \rightarrow 1$
& Same as VG-NPU \\
\cline{2-4}

& Enhanced quantization
& None $\rightarrow$ DXQ-P3
& Same as VG-NPU \\
\hline

\multirow{3}{*}[-1.5ex]{Regulus}
& optimize\_option
& $0 \rightarrow 1$
& Same as VG-NPU \\
\cline{2-4}

& Calibration percentile
& $0.999 \rightarrow 0.9999$
& Same as VG-NPU \\
\cline{2-4}

& Hadamard rotation
& No change
& Off $\rightarrow$ On \\
\hline

\end{tabular}

\vspace{2pt}
\begin{flushleft}
\footnotesize
W4 $\times 3$ and W4 $\times 4$ indicate that three and four layers, respectively, use 4-bit weights. W8A8 denotes 8-bit weights and 8-bit activations.
\end{flushleft}

\end{table}

For Hailo-8 and Hailo-10H, VG-NPU increases the optimization and compression levels and applies workload-specific mixed-precision configurations. On Hailo-8, three W4 layers are introduced for both ResNet50 and YOLOv5m, while YOLOv8s-pose uses A16/W16 precision for the keypoint-related outputs without W4 layers. On Hailo-10H, VG-NPU uses three W4 layers for ResNet50, four for YOLOv5m, and three for YOLOv8s-pose. Additional model-specific settings include quantization finetuning, bias correction, and an allocator adjustment for YOLOv5m. Under VG-E2E, some of these precision settings are further changed according to the vendor Model Zoo recipe.

For DX-M1, VG-NPU increases \texttt{opt\_level} from 0 to 1 and enables the enhanced DXQ-P3 quantization configuration. These settings are retained in VG-E2E.

For Regulus, VG-NPU increases \texttt{optimize\_option} from 0 to 1 and changes the calibration percentile from 0.999 to 0.9999. Hadamard rotation remains disabled in VG-NPU and is enabled in VG-E2E as part of the vendor-native configuration.

\section{Vendor-Specific Deployment Changes}
\label{app:deployment-changes}

Table~\ref{tab:vendor-deployment-changes} summarizes the deployment-level changes from the common Baseline/VG-NPU execution path to VG-E2E. Baseline and VG-NPU use the same model Input/Output, host-side preprocessing and postprocessing, and benchmark runtime pipeline, whereas VG-E2E follows the vendor-native deployment path.

\begin{table}[h]
\centering
\caption{Deployment configuration changes from the common Baseline/VG-NPU path to VG-E2E.}
\label{tab:vendor-deployment-changes}

\setlength{\tabcolsep}{3pt}
\renewcommand{\arraystretch}{1.15}
\footnotesize

\begin{tabular}{
|C{1.70cm}
|L{2.20cm}
|L{3.60cm}
|L{4.00cm}|}
\hline

\multicolumn{1}{|c|}{\textbf{NPU}} &
\multicolumn{1}{c|}{\textbf{Item}} &
\multicolumn{1}{c|}{\textbf{Baseline / VG-NPU}} &
\multicolumn{1}{c|}{\textbf{VG-E2E}} \\
\hline

\multirow[c]{6}{*}[-5ex]{\shortstack[c]{Hailo-8\\Hailo-10H}}
& Input
& Normalized FP32
& Raw RGB UINT8 \\
\cline{2-4}

& Preprocessing
& Host
& Vendor-native preprocessing \\
\cline{2-4}

& YOLOv5m output
& 3 detection heads
& NMS output $[80,5,80]$ \\
\cline{2-4}

& YOLOv8s-pose output
& 9 tensors for bbox, class, and keypoints
& 9 heads: 3 scales $\times$ \{64, 1, 51\} \\
\cline{2-4}

& Postprocessing
& Common postprocessor
& Hailo vendor postprocessor \\
\cline{2-4}

& Runtime / pipeline
& Benchmark pipeline
& TAPPAS (GStreamer) \\
\hline

\multirow[c]{6}{*}[-5ex]{DX-M1}
& Input
& Normalized FP32
& Raw RGB UINT8 \\
\cline{2-4}

& Preprocessing
& Host
& Vendor-native preprocessing \\
\cline{2-4}

& YOLOv5m output
& 3 detection heads
& Structured BBOX $[25200]$ \\
\cline{2-4}

& YOLOv8s-pose output
& 9 tensors for bbox, class, and keypoints
& output0 $[1,56,8400]$ \\
\cline{2-4}

& Postprocessing
& Common postprocessor
& DeepX vendor postprocessor \\
\cline{2-4}

& Runtime / pipeline
& Benchmark pipeline
& DXSTREAM (GStreamer) \\
\hline

\multirow[c]{6}{*}[-5ex]{Regulus}
& Input
& Normalized FP32
& Raw RGB UINT8 \\
\cline{2-4}

& Preprocessing
& Host
& Vendor-native preprocessing \\
\cline{2-4}

& YOLOv5m output
& 3 detection heads
& output0 $[1,25200,85]$ \\
\cline{2-4}

& YOLOv8s-pose output
& 9 tensors for bbox, class, and keypoints
& output0 $[1,56,8400]$ \\
\cline{2-4}

& Postprocessing
& Common postprocessor
& Mobilint Model Zoo postprocessor \\
\cline{2-4}

& Runtime / pipeline
& Benchmark pipeline
& \texttt{qbruntime} \\
\hline

\end{tabular}
\end{table}

In VG-E2E, preprocessing follows the vendor-native deployment path rather than the common host-side preprocessing used in Baseline and VG-NPU. For Hailo, image decoding and resizing are handled through the TAPPAS pipeline, which is based on GStreamer, while normalization is integrated into the compiled model. DeepX similarly performs image-level processing through DXSTREAM, which is based on GStreamer, and applies input scaling and normalization within the model. For Regulus, preprocessing is provided through the Mobilint Model Zoo, with scaling and normalization integrated into the compiled model.

The model output representation also differs across vendor-native deployments. For Hailo YOLOv5m, NMS is integrated into the compiled model execution path, resulting in an output of $[80,5,80]$. The Hailo pose model exposes 9 outputs over three scales, consisting of bbox, class, and keypoint components. For DX-M1 YOLOv5m, the PPU generates 25,200 structured bounding-box records, while YOLOv8s-pose produces a $[1,56,8400]$ output assembled through the model's CPU subgraph. Regulus produces packed outputs of $[1,25200,85]$ and $[1,56,8400]$ for YOLOv5m and YOLOv8s-pose, respectively.

Although some VG-E2E configurations expose identical or similar output shapes, the operations performed before producing each output differ across vendors. Therefore, these output representations should not be interpreted as indicating identical execution scopes or operator placement.

\end{document}